\documentclass[akbc,twoside,11pt,lettersize]{article}
\usepackage{akbc}
\usepackage{times}
\usepackage{latexsym}
\usepackage{microtype}
\usepackage{amsmath}
\usepackage{tikz}
\usepackage{pgfplots}
\usepackage{graphicx}
\usepackage{amssymb}
\usepackage{amsfonts}
\usepackage{amsthm}
\usepackage{mathtools,amssymb}
\usepackage{booktabs}
\usepackage{tablefootnote}
\usepackage{subcaption}
\usepackage{enumitem}
\usepackage{multirow}
\usepackage[ruled, linesnumbered, vlined]{algorithm2e}
\usepackage{verbatim}
\SetKwInput{KwInput}{Input}                
\SetKwInput{KwOutput}{Output}
\akbcheading
\ShortHeadings{Few-Shot Inductive Learning on TKGs using Concept-Aware Information}{Ding, Wu, He, Ma, Han \& Tresp}
\finalcopy 
\begin{document}

\title{Few-Shot Inductive Learning on Temporal Knowledge Graphs using Concept-Aware Information}

\author{\name Zifeng Ding\thanks{\;Equal contribution.}$\; ^{1,2}$ \email zifeng.ding@campus.lmu.de\\
       {\bf
       \name Jingpei Wu$^{*3}$ \email jingpei.wu@tum.de
       }\\
       {\bf
       \name Bailan He$^{1}$ \email bailan.he@campus.lmu.de
       }\\
       {\bf
       \name Yunpu Ma$^{1,2}$ \email cognitive.yunpu@gmail.com
       }\\
       {\bf
       \name Zhen Han\thanks{\;Corresponding author.} $^{1,2}$ \email zhen.han@campus.lmu.de
       }\\
       {\bf
       \name Volker Tresp$\footnotemark[2]$ $^{1,2}$ \email volker.tresp@siemens.com
       }\\
       $^{1}$\addr{Ludwig Maximilian University of Munich}\\
       $^{2}$\addr{Corporate Technology, Siemens AG}\\
       $^{3}$\addr{Technical University of Munich}
       }


\maketitle

\begin{abstract}
Knowledge graph completion (KGC) aims to predict the missing links among knowledge graph (KG) entities. Though various methods have been developed for KGC, most of them can only deal with the KG entities seen in the training set and cannot perform well in predicting links concerning novel entities in the test set. 
Similar problem exists in temporal knowledge graphs (TKGs), and no previous temporal knowledge graph completion (TKGC) method is developed for modeling newly-emerged entities. Compared to KGs, TKGs require temporal reasoning techniques for modeling, which naturally increases the difficulty in dealing with novel, yet unseen entities. In this work, we focus on the inductive learning of unseen entities' representations on TKGs. We propose a few-shot out-of-graph (OOG) link prediction task for TKGs, where we predict the missing entities from the links concerning unseen entities by employing a meta-learning framework and utilizing the meta-information provided by only few edges associated with each unseen entity. We construct three new datasets for TKG few-shot OOG link prediction, and we propose a model that mines the concept-aware information among entities. Experimental results show that our model achieves superior performance on all three datasets and our concept-aware modeling component demonstrates a strong effect.
\end{abstract}

\section{Introduction}
\label{Introduction}
Knowledge graphs (KGs) store factual information in the form of triples, i.e., $(s,r,o)$, where $s$, $o$, $r$ denote the subject entity, the object entity, and the relation between them, respectively. KGs have already been widely used in a series of downstream tasks, e.g., question answering \cite{saxena-etal-2020-improving,DBLP:journals/corr/abs-2208-06501} and recommender systems \cite{DBLP:conf/aaai/WangWX00C19,DBLP:conf/www/WangZXLG19}. While KG triples are capable of representing facts, they cannot express their time validity. World knowledge is ever-changing, which means many facts have their own time validity, e.g., the fact (\textit{Angela Merkel}, \textit{is chancellor of}, \textit{Germany}) is valid only before (\textit{Olaf Scholz}, \textit{is chancellor of}, \textit{Germany}). To this end, temporal knowledge graphs (TKGs) are introduced to consider the time validity of facts by representing every fact with a quadruple, i.e., $(s,r,o,t)$, where $t$ denotes the time when the fact is valid.

KGs and TKGs are known to suffer from incompleteness \cite{DBLP:conf/naacl/MinGW0G13,DBLP:conf/www/LeblayC18}. Therefore, various methods have been developed for automatically completing KGs \cite{DBLP:conf/icml/NickelTK11,DBLP:conf/nips/BordesUGWY13,DBLP:conf/icml/TrouillonWRGB16,DBLP:conf/iclr/SunDNT19,DBLP:conf/emnlp/GuoK21} and TKGs \cite{tresp2015learning,DBLP:conf/www/LeblayC18,Ma2019EmbeddingMF,DBLP:conf/kdd/JungJK21,DBLP:journals/corr/abs-2112-07791}. Though these methods achieve superior performance on knowledge graph completion (KGC) and temporal knowledge graph completion (TKGC), they have their limitations. In real-world scenarios, KGs and TKGs evolve over time, indicating that new (unseen) entities may emerge constantly \cite{DBLP:conf/aaai/ShiW18}. Besides, real-world KGs exhibit long-tail distributions, where a large portion of entities only have few edges \cite{DBLP:conf/nips/BaekLH20}. This also applies to TKGs, e.g., the entity frequency distribution of ICEWS datasets (Appendix \ref{app: long tail}). Traditional KGC and TKGC methods learn the representations of the observed (seen) entities, and perform link prediction over a fixed set of entities. To learn the optimal representations of the observed entities, these methods require a large number of training examples associated with each of them. \cite{DBLP:conf/nips/BaekLH20} shows that traditional KGC methods show poor performance when they are used to predict the links concerning newly-emerged, yet unseen entities. In our work, we also observe that traditional TKGC methods share the same problem (Section \ref{exp results}).

To tackle the limitations of traditional TKGC methods, we propose the TKG few-shot out-of-graph (OOG) link prediction task and a TKG reasoning model for better learning the inductive representations of newly-emerged entities in TKGs. Inspired by recent work that mines shared concepts of stocks for improving stock prediction \cite{DBLP:conf/ijcai/0101BHCXS20,DBLP:journals/corr/abs-2110-13716}, we devise a module, taking advantage of the entity concepts provided by the temporal knowledge bases. The contribution of our work is three-folded:
\begin{itemize}
    \item We propose the TKG few-shot out-of-graph (OOG) link prediction task. To better learn the inductive representations of unseen entities and predict their links, we propose a meta-learning-based model. To the best of our knowledge, this is the first work aiming to improve the link prediction performance concerning unseen entities in TKGs.
    \item We extract the entity concepts from the temporal knowledge bases and take them as additional information to boost our model performance. We design an effective module to learn concept-aware information. The experimental results show that introducing such information helps to learn better representations for unseen entities in the inductive setting.
    \item We propose three new datasets for TKG few-shot OOG link prediction, i.e., ICEWS14-OOG, ICEWS18-OOG and ICEWS0515-OOG. We compare our model with several baseline methods. Experimental results show that our model outperforms all the baselines on all three datasets.
\end{itemize}

\section{Related Work}
\paragraph{Knowledge graph embedding methods.}
Knowledge graph embedding (KGE) methods can be split into two categories. Some methods design scoring functions to compute the plausibility scores of KG facts \cite{DBLP:conf/nips/BordesUGWY13,DBLP:conf/icml/TrouillonWRGB16,DBLP:conf/iclr/SunDNT19,DBLP:conf/emnlp/GuoK21}, while other KGE methods employ neural-based structures, e.g., graph neural networks (GNNs), to better capture the structural dependencies of KGs \cite{DBLP:conf/esws/SchlichtkrullKB18,DBLP:conf/iclr/VashishthSNT20,DBLP:conf/www/YuYZW21}. By combining neural-based graph encoders with KG scoring functions, these methods achieve superior performance in KG reasoning tasks.

\paragraph{Temporal knowledge graph embedding methods.}
To deal with the temporal constraints in TKG facts, two lines of temporal knowledge graph embedding (TKGE) methods have been developed. The first line of methods designs novel time-aware scoring functions for characterizing extra time information \cite{DBLP:conf/www/LeblayC18,Ma2019EmbeddingMF,DBLP:conf/iclr/LacroixOU20,DBLP:conf/aaai/SadeghianACW21,han2020dyernie, han-etal-2021-time}. The second line of methods models temporal information by employing neural structures, e.g., GNNs and recurrent models. \cite{DBLP:conf/iclr/HanCMT21,DBLP:conf/kdd/JungJK21,DBLP:journals/corr/abs-2112-07791, han2020graph, sun-etal-2021-timetraveler} sample every entity's temporal neighbors and use GNNs to learn time-aware representations of them. \cite{DBLP:conf/emnlp/WuCCH20} and \cite{DBLP:conf/emnlp/HanDMGT21} model structural information with GNNs, and they achieve temporal reasoning by utilizing a gated recurrent unit \cite{DBLP:conf/emnlp/ChoMGBBSB14} and a neural ordinary differential equation \cite{DBLP:conf/nips/ChenRBD18}, respectively.

\paragraph{Inductive learning on knowledge graphs.}
Traditional KGE and TKGE methods require a large number of training examples to learn entity representations. However, in real-world scenarios, KGs and TKGs are ever-evolving, and they exhibit long-tail distributions. New entities and relations emerge and a huge portion of them only have very few associated facts, thus causing traditional methods unable to learn optimal representations. To alleviate this problem, a line of work \cite{DBLP:conf/emnlp/XiongYCGW18,DBLP:conf/emnlp/ChenZZCC19,DBLP:conf/emnlp/ShengGCYWLX20,mirtaheri2021one,DBLP:journals/corr/abs-2205-10621} tries to employ meta-learning to learn inductive representations of unseen KG (or TKG) relations. Nevertheless, they are unable to deal with novel entities. Several methods try to deal with unseen (out-of-graph) entities in an inductive setting \cite{DBLP:conf/ijcai/HamaguchiOSM17,DBLP:conf/aaai/WangHLP19,DBLP:conf/cikm/HeWZTR20}. They first learn representations of seen entities, and then use an auxiliary set to transfer knowledge from seen to unseen entities during inference. \cite{DBLP:conf/nips/BaekLH20} proposes a more realistic task: few-shot out-of-graph (OOG) link prediction, where the links among unseen entities are also considered during evaluation and the representation of every unseen entity can only be derived from very few (number of shot size) edges. Baek et al. simulate the unseen entities in the training phase and introduce meta-learning for learning unseen entities' representations. Based on it, \cite{DBLP:conf/cikm/ZhangW0XLZ21} proposes a model using hyper-relation features to improve performance on few-shot OOG link prediction. Another series of work tries to include external information of entities, e.g., textual descriptions, to solve this problem \cite{DBLP:conf/aaai/XieLJLS16,DBLP:conf/emnlp/WangLLBL19} and it turns out to be effective in modeling unseen entities. Though there exist various methods dealing with OOG unseen entities in KGs, there is still no method specifically designed to embed unseen entities inductively for TKGs.

\section{Preliminaries and Task Formulation}
\paragraph{Entity concepts in temporal knowledge graphs.}
Entity concepts describe the characteristics of KG entities. They are manually defined by humans and assigned to every KG entity. In the ICEWS database \cite{DVN/28075_2015}, entities belong to several sectors, e.g., \textit{Government}, \textit{Executive Office}. Each entity's sectors are specified in the ICEWS weekly event data\footnote{https://dataverse.harvard.edu/dataset.xhtml?persistentId=doi:10.7910/DVN/QI2T9A}. We treat the sectors of an entity as its concepts and learn concept representations as additional information. We observe that some region entities in the ICEWS database, e.g., South Korea and North America, have no specified sectors. We manually assign a new sector \textit{Region} to them. We ensure that every entity has its own sectors. More details about concept extraction is presented in Appendix \ref{app: concept extract}.

\paragraph{Task formulation.}
We first give the definition of a temporal knowledge graph, then we formulate the TKG few-shot out-of-graph link prediction task.

\noindent
\textbf{Definition 1 (Temporal Knowledge Graph (TKG)).} Let $\mathcal{E}$, $\mathcal{R}$ and $\mathcal{T}$ denote a finite set of entities, relations and timestamps, respectively. A temporal knowledge graph (TKG) $\mathcal{G}$ can be taken as a finite set of TKG facts represented by their associated quadruples, i.e., $\mathcal{G} = \{(s,r,o,t)|s,o \in \mathcal{E}, r \in \mathcal{R}, t \in \mathcal{T}\} \subseteq \mathcal{E} \times \mathcal{R} \times \mathcal{E} \times \mathcal{T}$.


\noindent
\textbf{Definition 2 (Temporal Knowledge Graph Few-Shot Out-of-Graph Link Prediction).} Given an observed background TKG $\mathcal{G}_{\text{back}} \subseteq \mathcal{E}_{\text{back}} \times \mathcal{R} \times \mathcal{E}_{\text{back}} \times \mathcal{T}$, an unseen entity $e'$ is an entity $e' \in \mathcal{E}'$, where $\mathcal{E}' \cap \mathcal{E}_{\text{back}} = \emptyset$. Assume we further observe $K$ associated quadruples for each unseen entity $e'$ in the form of $(e', r, \Tilde{e}, t)$ (or $(\Tilde{e}, r, e', t)$), where $\Tilde{e} \in (\mathcal{E}_{\text{back}} \cup \mathcal{E}')$, $r \in \mathcal{R}$, $t \in \mathcal{T}$, and $K$ is a small number denoting the shot size, e.g., 1 or 3. TKG few-shot out-of-graph link prediction aims to predict the missing entities from the link prediction queries $(e', r_q, ?, t_q)$ (or $(?, r_q, e', t_q)$) derived from unobserved quadruples containing unseen entities, where $r_q \in \mathcal{R}$, $t_q \in \mathcal{T}$.

We further formulate the TKG few-shot OOG link prediction task into a meta-learning problem. For a TKG $\mathcal{G} \subseteq \mathcal{E} \times \mathcal{R} \times \mathcal{E} \times \mathcal{T}$, we first select a group of entities $\mathcal{E}'$, where each entity's number of associated quadruples is between a lower and a higher threshold. We aim to pick out the entities that are not frequently mentioned in TKG facts since newly-emerged entities normally are coupled with only several edges. We randomly split these entities into three groups $\mathcal{E}'_{\text{meta-train}}$, $\mathcal{E}'_{\text{meta-valid}}$ and $\mathcal{E}'_{\text{meta-test}}$. For each group, we treat the union of all the quadruples associated to this group's entities as the corresponding meta-learning set, e.g., the meta-training set $\mathbb{T}_{\text{meta-train}}$ is formulated as $\{(e',r, \Tilde{e},t)|\Tilde{e} \in \mathcal{E}, r \in \mathcal{R}, e' \in \mathcal{E}'_{\text{meta-train}}, t \in \mathcal{T}\} \cup \{(\Tilde{e},r, e',t)|\Tilde{e} \in \mathcal{E}, r \in \mathcal{R}, e' \in \mathcal{E}'_{\text{meta-train}}, t \in \mathcal{T}\}$. We ensure that there exists no link between every two of the meta-learning sets. The associated quadruples of the rest entities form a background graph $\mathcal{G}_{\text{back}} \subseteq \mathcal{E}_{\text{back}} \times \mathcal{R} \times \mathcal{E}_{\text{back}} \times \mathcal{T}$, where $\mathcal{E}' \cap \mathcal{E}_{\text{back}} = \emptyset$ and $\mathcal{E} = (\mathcal{E}_{\text{back}} \cup \mathcal{E}')$. We take the meta-training entities $\mathcal{E}'_{\text{meta-train}}$ as simulated unseen entities and try to learn how to transfer knowledge from seen entities $\mathcal{E}_{\text{back}}$ to them during meta-training. The entities in $\mathcal{E}'_{\text{meta-valid}}$ and $\mathcal{E}'_{\text{meta-test}}$ are real unseen entities that are used to evaluate the model performance.

Based on \cite{DBLP:conf/nips/BaekLH20}, we define a meta-training task $T$ as follows. In each task $T$, we first randomly sample $N$ simulated unseen entities $\mathcal{E}_{T}$ from $\mathcal{E}'_{\text{meta-train}}$. Then we randomly select $K$ associated quadruples for each $e' \in \mathcal{E}_{T}$ as its support quadruples $\mathcal{S}_{e'} = \{(e',r_i,\Tilde{e}_i,t_i) \ \text{or}\  (\Tilde{e}_i,r_i,e',t_i)\}^K_{i=1}$, where $K$ is the shot size and $\Tilde{e}_i \in (\mathcal{E}_{\text{back}} \cup \mathcal{E}')$. The rest of $e'$'s quadruples are taken as its query quadruples $\mathcal{Q}_{e'} = \{(e',r_i,\Tilde{e}_i,t_i) \ \text{or}\  (\Tilde{e}_i,r_i,e',t_i)\}^{M_{e'}}_{i=K+1}$, where $M_{e'}$ denotes the number of $e'$'s associated quadruples in $\mathbb{T}_{\text{meta-train}}$ and $\Tilde{e}_i \in (\mathcal{E}_{\text{back}} \cup \mathcal{E}')$. For every meta-training task $T$, the aim of TKG few-shot OOG link prediction is to simultaneously predict the missing entities from the link prediction queries derived from the query quadruples associated to all the entities from $\mathcal{E}_T$, e.g., $(e', r_i, ?, t_i)$ or $(?, r_i, e', t_i)$. In this way, we simulate the situation that we simultaneously observe a bunch of unseen entities and each of them has only few edges, which is similar to how emerging entities appear in temporal knowledge bases. After meta-training, we validate our model on a meta-validation set $\mathbb{T}_{\text{meta-valid}}$ and test our model on a meta-test set $\mathbb{T}_{\text{meta-test}}$, where they contain all the quadruples associated to the entities in $\mathcal{E}'_{\text{meta-valid}}$ and $\mathcal{E}'_{\text{meta-test}}$, respectively. We do not sample $N$ entities during meta-validation and meta-test. Instead, we treat all the entities in $\mathcal{E}'_{\text{meta-valid}}$ (or $\mathcal{E}'_{\text{meta-test}}$) as appearing at the same time. For a better understanding, we present Figure \ref{fig: meta-learning} to illustrate how we formulate the TKG few-shot OOG link prediction task into a meta-learning problem. We also discuss the difference between our proposed task and traditional TKGC in Appendix \ref{app: oog link prediction}.

We summarize the challenge of TKG few-shot OOG link prediction as follows: (1) TKG reasoning models are asked to predict the links concerning the newly-emerged entities that are completely unseen during the training process; (2) Only a small number ($K$) of edges associated with each newly-emerged entity are observable to support predicting the unobserved links concerning this entity.

\section{Our Method}
\begin{figure}[htbp]
\centering
\includegraphics[width=0.9\columnwidth]{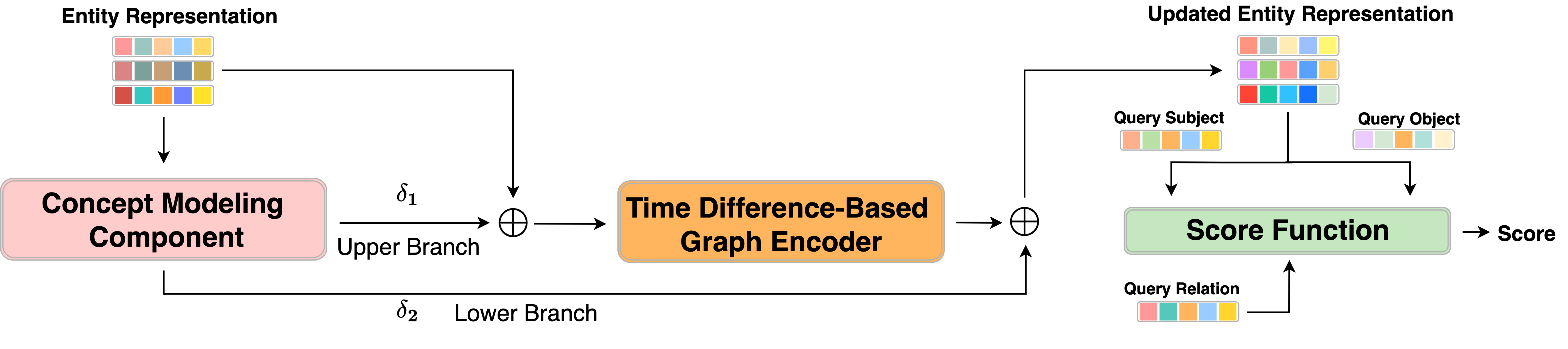}
\caption{\label{fig: model structure} Model structure of FILT. Assume we have an unseen entity $e'$, and we want to predict a link corresponding to $(e', r_i, \Tilde{e}_i, t_i) \in \mathcal{Q}_{e'}$. We derive the concept representations in the concept modeling component and use a time difference-based graph encoder for learning $e'$'s time-aware representation. We take the representations of $r_i$ and $\Tilde{e}_i$ to compute the plausibility score of the link.}
\end{figure}
\noindent
We propose a model dealing with \textbf{f}ew-shot \textbf{i}nductive \textbf{l}earning on \textbf{T}KGs (FILT). Figure \ref{fig: model structure} shows the model structure of FILT. It consists of three components: (1) Concept modeling component that represents entity concepts based on seen entities' representations; (2) Time difference-based graph encoder that learns the contextualized representations of unseen entities; (3) KG scoring function that computes the plausibility scores of the TKG quadruples concerning unseen entities.
\subsection{Concept Modeling Component}
When a new entity emerges in a TKG, though there might be only few observed associated edges, some of its concepts, e.g., which sectors it belongs to, are already known. Since every entity concept is shared across all the entities in this TKG, we can learn concept information from seen entities and transfer it to newly-emerged entities. 

Inspired by \cite{DBLP:journals/corr/abs-2110-13716} that mines concept-aware information for stock prediction, we develop a concept modeling component to learn TKG entity concepts as follows. First, we pre-train our background graph with ComplEx \cite{DBLP:conf/icml/TrouillonWRGB16}. Note that only seen entities $\mathcal{E}_{\text{back}}$ are involved in the pre-training process. Assume we have a set of entity concepts $\mathcal{C}$, then we initialize the representation of every entity concept $c \in \mathcal{C}$ with its associated entities by averaging these entities' representations:
\begin{equation}
    \mathbf{h}_{c} = \frac{1}{|\mathcal{N}_c|}\sum_{e \in \mathcal{N}_c} \mathbf{h}_{e},
\end{equation}
where $\mathbf{h}_c$ and $\mathbf{h}_e$ denote the representations of the concept $c$ and the entity $e$, respectively. $\mathcal{N}_c$ denotes the neighborhood of the entity concept $c$. For example, if two TKG entities \textit{Angela  Merkel} and \textit{Xi Jinping} both belong to the concept \textit{Elite}, they will be included into \textit{Elite}'s neighborhood. Since we want to distinguish the contributions of different entities to an entity concept, we then correct the concept representations as follows:
\begin{equation}
\label{eq: correct concept}
     \mathbf{h}_{c} =  \sum_{e_i \in \mathcal{N}_c} \alpha_c^{e_i} \mathbf{h}_{e_i}, \quad\alpha_c^{e_i} = \frac{\exp(\mathbf{h}^{\top}_{e_i}\mathbf{h}_{c})}{\sum_{e_j \in \mathcal{N}_c}\exp(\mathbf{h}^{\top}_{e_j}\mathbf{h}_c)}.
\end{equation}
After we correct the concept representations, we compute an entity's concept-aware information by aggregating the representations of its associated concepts:
\begin{equation}
\label{eq: concept agg}
\begin{aligned}
     \mathbf{h}_{e}^{\mathcal{C}_e} =  \sum_{c_i \in \mathcal{C}_e} \beta_e^{c_i} \mathbf{h}_{c_i}, \quad\beta_e^{c_i} = \frac{\exp(\mathbf{h}^{\top}_{c_i}\mathbf{h}_{e})}{\sum_{c_j \in \mathcal{C}_e}\exp(\mathbf{h}^{\top}_{c_j}\mathbf{h}_e)}.
\end{aligned}
\end{equation}
$\mathcal{C}_e \subseteq \mathcal{C}$ denotes the set of all concepts associated to $e$. As shown in Figure \ref{fig: model structure}, we inject the concept-aware information into two branches. We use two separate layers of feed forward neural network and project the concept-aware information onto two branches. The upper branch adds the concept information to the entity representations $\mathbf{h}_e := \mathbf{h}_e + \delta_1 \sigma(\mathbf{W}_c^1\mathbf{h}_{e}^{\mathcal{C}_e})$ and take them as the input of our graph encoder. The lower branch processes the concept information $\delta_2 \sigma(\mathbf{W}_c^2\mathbf{h}_{e}^{\mathcal{C}_e})$ and adds it to the entity representations after the graph aggregation step. $\delta_1$ and $\delta_2$ are two trainable weights deciding how much concept-aware information should be injected. $\mathbf{W}_c^1$ and $\mathbf{W}_c^2$ are two weight matrices and $\sigma$ is an activation function. By employing the double branch structure, we not only include the concept information into the graph encoder, but also directly infuse it into the final entity representations for link prediction.
\subsection{Time Difference-Based Graph Encoder}
\begin{figure}[htbp]
\centering
\includegraphics[scale=0.45]{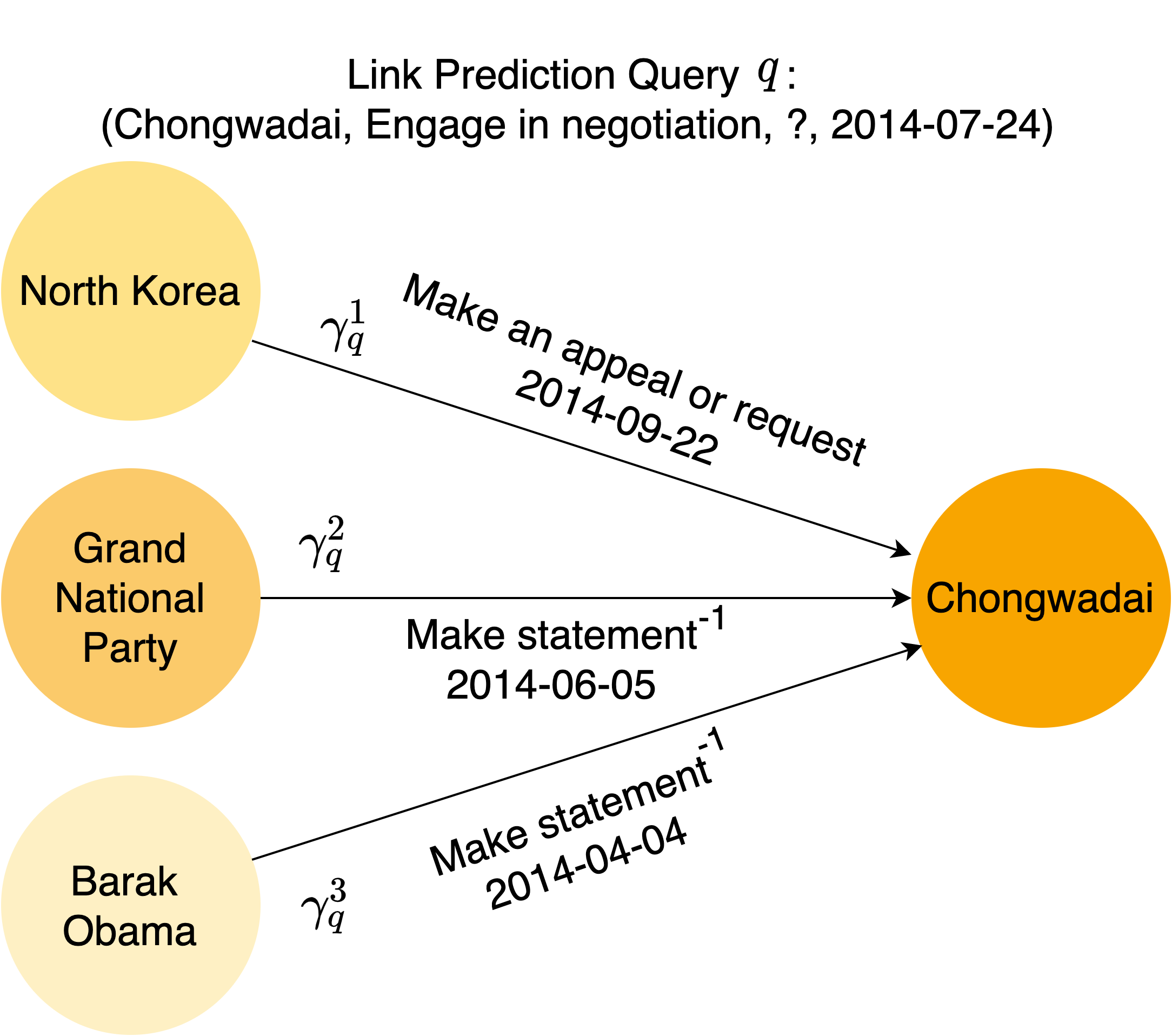}
\caption{\label{fig: encoder structure} The structure of the time difference-based graph encoder. Assume we have an unseen entity \textit{Chongwadai}, and we have a link prediction query (\textit{Chongwadai}$ $, \textit{Engage in negotiation}, $?$, \textit{2014-07-24}), given three support quadruples, i.e., (\textit{North Korea}, \textit{Make an appeal or request}, \textit{Chongwadai}, \textit{2014-09-22}), (\textit{Chongwadai}, \textit{Make statement}, \textit{Grand National Party}, \textit{2014-06-05}), and (\textit{Chongwadai}, \textit{Make statement}, \textit{Barak Obama}, \textit{2014-04-04}). We use our graph encoder to compute the time-aware contextualized representation of \textit{Chongwadai} at \textit{2014-07-24}. For each temporal neighbor from a support quadruple, we compute its importance according to the time difference between \textit{2014-07-24} and the timestamp of its corresponding support quadruple. We denote the temporal neighbors with colored circles. The color darkness of the circles implies the importance of the temporal neighbors during aggregation in Equation \ref{eq: encoder agg}. The darker circle a temporal neighbor is represented with, the more important it is, i.e., $\gamma_{q}^2 > \gamma_{q}^1 > \gamma_{q}^3$.}
\end{figure}

\noindent
To compute the contextualized representations of the unseen entities, we employ a time difference-based graph encoder. For each unseen entity $e'$, assume we have a link prediction query $(e', r_q, ?, t_q)$ derived from a query quadruple $(e', r_q, \Tilde{e}_q, t_q)\in \mathcal{Q}_{e'}$. We first find its temporal neighbors from its support quadruples $\mathcal{S}_{e'} = \{(e',r_i,\Tilde{e}_i,t_i) \ \text{or}\  (\Tilde{e}_i,r_i,e',t_i)\}^K_{i=1}$, and then compute $e'$'s time-aware representation at $t_q$ through aggregation:
\begin{equation}
\label{eq: encoder agg}
\begin{aligned}
     \mathbf{h}_{(e', t_q)} = \sum_{(\Tilde{e}_i, r_i, t_i) \in \mathcal{N}_{e'}} \gamma_{q}^i  \mathbf{W_g}(\mathbf{h}_{\Tilde{e}_i} \|\mathbf{h}_{r_i}), \quad
     \gamma_{q}^i = \frac{\exp(1 / {|t_q - t_i|})}{\sum_{(\Tilde{e}_j, r_j, t_j)\in \mathcal{N}_{e'}} \exp (1 / |t_q - t_j|)}.
\end{aligned}
\end{equation}
$\mathbf{W}_g$ denotes the weight matrix in our graph encoder. $\mathcal{N}_{e'}$ denotes the observed neighborhood of $e'$ and $|\mathcal{N}_{e'}| = K$. $\gamma_{q}^i$ is the importance of the $i$th temporal neighbor $\Tilde{e}_i$ based on the time difference between $t_q$ and $t_i$. The smaller the time difference is, the more important a temporal neighbor is during aggregation. The motivation of our time difference-based graph encoder is that we assume the temporal neighbors that are temporally closer to the query timestamp $t_q$ tend to contribute more to predicting the links at $t_q$. Since we take the temporal neighbors of an entity from its incoming edges, we transform every support quadruple whose form is $(e', r_i, \Tilde{e}_i, t_i)$ to $(\Tilde{e}_i, r^{-1}_i, e', t_i)$, where $r^{-1}_i$ corresponds to the inverse relation of $r_i$. We manage to incorporate every support quadruple into the aggregation process with this quadruple transformation. Note that if $t_q - t_i = 0$, the denominator of the exponential term will be $0$. Thus, we use a constant $\lambda$ to assign a value to $\exp(1 / {|t_q - t_i|})$ if $t_q$ equals $t_i$, and $\lambda$ serves as a hyperparameter that can be tuned. Figure \ref{fig: encoder structure} illustrates the structure of our graph encoder with an example. After aggregation, we further infuse the concept-aware information from the lower branch into the output of our graph encoder: $\mathbf{h}_{(e',t_q)} := \mathbf{h}_{(e',t_q)} + \delta_2 \sigma(\mathbf{W}_c^2\mathbf{h}_{e'}^{\mathcal{C}_{e'}})$. We show in Section \ref{ablation} that our simple-structured graph encoder can beat more complicated structures in the TKG OOG link prediction task.

\subsection{Parameter Learning}
For each meta-training task $T$, we have $N$ simulated unseen entities $\mathcal{E}_T$. We use the hinge loss for learning model parameters:
\begin{equation}
\label{eq: loss}
    \mathcal{L} = \sum_{e' \in \mathcal{E}_T} \sum_{q^+ \in \mathcal{Q}_{e'}} \sum_{q^- \in \mathcal{Q}^-_{e'}} 
    \text{max} \{\theta - score(q^+) + score(q^-),0\}.
\end{equation}
$\theta > 0$ is the margin. $q^+$ denotes a query quadruple from $e'$'s query set. $q^-$ is generated by negative sampling \cite{DBLP:conf/nips/BordesUGWY13}. For every $q^+ = (e', r_q, \Tilde{e}_q, t_q)$ (or $q^+ = (\Tilde{e}_q, r_q, e', t_q)$), we corrupt $\Tilde{e}_q$ with another entity $e^- \in \{\mathcal{E}', \mathcal{E}_{\text{back}}\}$. We map our learned representations to the complex space and use ComplEx \cite{DBLP:conf/icml/TrouillonWRGB16} as our scoring function, i.e., $score = \text{Re}<\mathbf{h}_s,\mathbf{h}_r, \Bar{\mathbf{h}}_o>$, where $\mathbf{h}_s$, $\mathbf{h}_o$ denote the representations of the subject entity and the object entity, respectively. $\mathbf{h}_r$ denotes the relation representation. $\text{Re}$ means taking the real part, and $\Bar{\mathbf{h}}_o$ means taking the conjugate of the vector $\mathbf{h}_o$.

\section{Experiments}
\label{results}
We compare FILT with several baselines on TKG few-shot OOG link prediction. To prove the effectiveness of the model components, we conduct several ablation studies. We also do further analysis to show the robustness of our method. Besides, we visualize the learned concept representations and show that our concept modeling component helps to capture the semantics of entity concepts.
\subsection{Datasets}
We propose three TKG few-shot OOG link prediction datasets, i.e., ICEWS14-OOG, ICEWS18-OOG, and ICEWS0515-OOG. We first take three subsets, i.e., ICEWS14, ICEWS18, and ICEWS05-15, from the Integrated Crisis Early Warning System (ICEWS) database \cite{DVN/28075_2015}, where they contain the timestamped political facts in 2014, in 2018, and from 2005 to 2015, respectively. Following the data construction process of \cite{DBLP:conf/nips/BaekLH20}, for each subset, we first randomly sample half of the entities whose number of associated quadruples is between a lower and a higher threshold as unseen entities. Then we split the sampled entities into three groups $\mathcal{E}'_{\text{meta-train}}$, $\mathcal{E}'_{\text{meta-valid}}$, $\mathcal{E}'_{\text{meta-test}}$ ($\mathcal{E}'_{\text{meta-train}} \cap \mathcal{E}'_{\text{meta-valid}} = \emptyset$, $\mathcal{E}'_{\text{meta-train}} \cap \mathcal{E}'_{\text{meta-test}} = \emptyset$, $\mathcal{E}'_{\text{meta-valid}} \cap \mathcal{E}'_{\text{meta-test}} = \emptyset$), where $|\mathcal{E}'_{\text{meta-train}}|:|\mathcal{E}'_{\text{meta-valid}}|:|\mathcal{E}'_{\text{meta-test}}| \approx 8:1:1$. The associated quadruples of all the entities in $\mathcal{E}'_{\text{meta-train}}$/$\mathcal{E}'_{\text{meta-valid}}$/$\mathcal{E}'_{\text{meta-test}}$ form the meta-training/meta-validation/meta-test set. The rest of the quadruples without unseen entities are used for constructing a background graph $\mathcal{G}_{\text{back}}$. The dataset statistics are presented in Table \ref{tab: data}. We present the dataset construction process in Appendix \ref{app: data construct}.
\begin{table}[htbp]
    \centering
    \resizebox{0.9\columnwidth}{!}{
\begin{tabular}{c c c c c c c c c c c} \hline
\textbf{Dataset}&$|\mathcal{E}|$&$|\mathcal{R}|$&$|\mathcal{T}|$&$|\mathcal{E}'_{\text{meta-train}}|$&$|\mathcal{E}'_{\text{meta-valid}}|$&$|\mathcal{E}'_{\text{meta-test}}|$&$N_{\text{back}}$&$N_{\text{meta-train}}$&$N_{\text{meta-valid}}$&$N_{\text{meta-test}}$\\ 
\hline
ICEWS14-OOG  & 7128 & 230 & 365 & 385 & 48 & 49 & 83448 & 5772 & 718 & 705 \\ 
ICEWS18-OOG  & 23033 & 256 & 304 & 1268 & 160 & 158 & 444269 & 19291 & 2425 & 2373\\ 
ICEWS0515-OOG  & 10488 & 251 & 4017 & 647 & 80 & 82 & 448695 & 10115 & 1217 & 1228\\ 
\hline
\end{tabular}
}
\caption{Dataset statistics. $|\mathcal{E}'_{\text{meta-train}}|$, $|\mathcal{E}'_{\text{meta-valid}}|$, $|\mathcal{E}'_{\text{meta-test}}|$ denote the number of unseen entities in the meta-training set, meta-validation set, meta-test set, respectively. $N_{\text{back}}$ denotes the number of quadruples in the background graph $\mathcal{G}_{\text{back}}$. $N_{\text{meta-train}}$, $N_{\text{meta-valid}}$, $N_{\text{meta-test}}$ denote the number of quadruples concerning unseen entities in $\mathbb{T}_{\text{meta-train}}$, $\mathbb{T}_{\text{meta-valid}}$, $\mathbb{T}_{\text{meta-test}}$, respectively.}
\label{tab: data}
\end{table}


\subsection{Baseline Methods}
We take four types of methods as our baselines. First we consider two traditional KGC methods, i.e., ComplEx \cite{DBLP:conf/icml/TrouillonWRGB16} and BiQUE \cite{DBLP:conf/emnlp/GuoK21}. Then we consider several traditional TKGC methods, i.e., TNTComplEx \cite{DBLP:conf/iclr/LacroixOU20}, TeLM \cite{DBLP:conf/naacl/XuCNL21}, and TeRo \cite{DBLP:conf/coling/XuNAYL20}. We combine all the quadruples in the background graph $\mathcal{G}_{\text{back}}$ with the quadruples of the meta-training set to construct a training set for traditional KGC as well as TKGC methods, and let them evaluate on all the query quadruples in the meta-validation/meta-test set. We also include two inductive KGC methods for OOG link prediction that do not employ meta-learning framework, i.e., MEAN \cite{DBLP:conf/ijcai/HamaguchiOSM17}, LAN \cite{DBLP:conf/aaai/WangHLP19}. To achieve fair comparison, we only allow them to utilize support quadruples during inference, rather than an auxiliary set containing a large number of quadruples for each unseen entity $e' \in \{\mathcal{E}'_{\text{meta-valid}},\mathcal{E}'_{\text{meta-test}}\}$. Apart from the first three types of methods, we further consider a meta-learning-based method GEN \cite{DBLP:conf/nips/BaekLH20} which deals with few-shot OOG link prediction on static KGs. For the baseline methods designed for static KGs, we provide them with all the quadruples in our datasets and neglect time constraints, i.e., neglecting $t$ in $(s,r,o,t)$. We ensure that all the methods evaluate exactly the same quadruples. 

\subsection{Experimental Results}
\label{exp results}
We report the TKG 1-shot and 3-shot OOG link prediction results in Table \ref{tab: main results}. We use mean reciprocal rank (MRR) and Hits@1/3/10 as the evaluation metrics (definition in Appendix \ref{app: eval metrics}). We follow the filtered setting \cite{DBLP:conf/nips/BordesUGWY13} for fairer evaluation. We observe that traditional KGC and TKGC methods show inferior performance in predicting the links concerning unseen entities. This is due to their nature that they have no way to transfer knowledge from seen to unseen entities. Besides, they learn representations of seen entities with a large number of associated training examples, thus causing the learned representations more prone to the data concerning seen entities and failing to embed unseen entities inductively. We also observe that inductive learning methods for static KGs show degenerated performance. MEAN, LAN, heavily rely on the auxiliary set during inference. We constrain their auxiliary set to only include the support quadruples, where only 1 associated quadruple for each unseen entity is included in the 1-shot case (3 associated quadruples in the 3-shot case). Experimental results show that these methods cannot effectively deal with newly-emerged entities that have only few observed edges, which is common in real-world scenarios. GEN employs meta-learning during training, thus having the ability to alleviate the data sparsity problem. However, it has no component to model temporal information, and it also does not incorporate any additional information, e.g., textual information and concept-aware information. To this end, GEN underperforms FILT in both 1-shot and 3-shot cases. Another crucial point worth noting is that the margin between FILT and GEN is much larger in the 3-shot case than in the 1-shot case. We attribute this to our time difference-based graph encoder. Our encoder distinguishes the importance of multiple support quadruples and aggregates the temporal neighbors more effectively.
\begin{table*}[htbp]
    \centering
    \resizebox{\columnwidth}{!}{
    \begin{tabular}{@{}lcccc cccc cccc cccc cccc cccc@{}}
\toprule
        \textbf{Datasets} & \multicolumn{8}{c}{\textbf{ICEWS14-OOG}} & \multicolumn{8}{c}{\textbf{ICEWS18-OOG}} & \multicolumn{8}{c}{\textbf{ICEWS0515-OOG}}\\
\cmidrule(lr){2-9} \cmidrule(lr){10-17} \cmidrule(lr){18-25}
& \multicolumn{2}{c}{MRR} & \multicolumn{2}{c}{H@1} & \multicolumn{2}{c}{H@3} & \multicolumn{2}{c}{H@10}& \multicolumn{2}{c}{MRR} & \multicolumn{2}{c}{H@1} & \multicolumn{2}{c}{H@3} & \multicolumn{2}{c}{H@10} & \multicolumn{2}{c}{MRR} & \multicolumn{2}{c}{H@1} & \multicolumn{2}{c}{H@3} & \multicolumn{2}{c}{H@10}\\

 \textbf{Model}& 1-S & 3-S & 1-S & 3-S & 1-S & 3-S & 1-S & 3-S & 1-S & 3-S & 1-S & 3-S & 1-S & 3-S & 1-S & 3-S & 1-S & 3-S & 1-S & 3-S & 1-S & 3-S & 1-S & 3-S\\
\midrule
        ComplEx & .048 & .046& .018 & .014&.045 & .046&.099 & .089
        & .039 & .044& .031 & .026&.048 & .042& .085 &.093
        & .077 & .076&.045 & .048&.074 & .071&.129 &.120
         \\
        BiQUE & .039 & .035& .015 & .014& .041 & .030& .073 &.066
        & .029 & .032& .022 & .021& .033 & .037& .064 &.073
        & .075 & .083&.044 & .049&.072 & .077&.130 &.144
         \\
\midrule 
        TNTComplEx & .043 & .044&.015 & .016&.033 & .042&.102&.096 
        & .046 & .048&.023 & .026&.043 & .044&.087& .082
        & .034 & .037&.014 & .012&.031 & .036&.060& .071
         \\
        TeLM & .032 & .035&.012 & .009&.021 & .023&.063 &.077
        & .049 & .019&.029 & .001&.045 & .013&.084&.054
        & .080 & .072&.041 & .034&.077 & .072&.138&.151
         \\
        TeRo  & .009 & .010&.002 & .002&.005 & .002&.015&.020 
        & .007 & .006&.003 & .001&.006 & .003&.013&.006
        & .012 & .023&.000 & .010&.008 & .017&.024&.040
         \\
        
\midrule
        MEAN  & .035 & .144&.013 & .054&.032 & .145&.082&.339 
        & .016 & .101&.003 & .014&.012 & .114&.043&.283
        & .019 & .148&.003 & .039&.017 & .175&.052&.384
       \\
       LAN  & .168 & .199&.050 & .061&.199 & .255&\textbf{.421}&\textbf{.500} 
        & .077 & .127&.018 & .025&.067 & .165&.199&.344
        & .171 & .182&.081 & .068&.180 & .191&.367&.467
       \\
\midrule     
      GEN  & .231 & .234&.162 & .155&.250 & .284&.378&.389 
        & .171 & .216&.112 & .137&.189 & .252&.289&.351
        & .268 & .322&.185 & .231&\textbf{.308} & .362&\textbf{.413}&.507
        \\       
\midrule
        FILT 
        & \textbf{.278} & \textbf{.321}&\textbf{.208} & \textbf{.240}&\textbf{.305} &\textbf{.357} &.410&.475
        & \textbf{.191} & \textbf{.266}&\textbf{.129} & \textbf{.187}&\textbf{.209} & \textbf{.298}&\textbf{.316}&\textbf{.417}
        & \textbf{.273} & \textbf{.370}&\textbf{.201} & \textbf{.299}&.303 & \textbf{.391}&.405&\textbf{.516}
        \\
\bottomrule
    \end{tabular}
    }
    \caption{TKG 1-shot and 3-shot OOG link prediction results. Evaluation metrics are filtered MRR and Hits@1/3/10 (H@1/3/10).
    The best results are marked in bold.}\label{tab: main results}
\end{table*}

\subsection{Ablation Study}
\label{ablation}
To prove the effectiveness of the model components, we conduct several ablation studies on ICEWS14-OOG and ICEWS18-OOG. We devise model variants in the following way. \textbf{(A) Concept Modeling Variants:} In A1 we run our model without the concept modeling component. In A2, we delete the lower branch connecting the concept modeling component with the output of the graph encoder. In A3, we delete the upper branch connecting the concept modeling component with the input of the graph encoder. \textbf{(B) Graph Encoder Variants:} In B1, we neglect the time information and switch our graph encoder to RGCN \cite{DBLP:conf/esws/SchlichtkrullKB18}. In B2, we use Time2Vec \cite{DBLP:journals/corr/abs-1907-05321} to model temporal information. In B3, we employ the functional time encoder introduced in \cite{DBLP:conf/iclr/XuRKKA20} as our graph encoder. In B4, we derive a time-aware attentional network as our graph encoder: $\mathbf{h}_{(e', t_q)} = \sum_{(\Tilde{e}_i, r_i, t_i) \in \mathcal{N}_{e'}} \gamma_q^i  \mathbf{W}_g(\mathbf{h}_{\Tilde{e}_i} \|\mathbf{h}_{r_i})$, where $\gamma_q^i = \frac{\exp \left( \sigma \left( ([\mathbf{h}_{r_q} \| \Phi(t_q)] \mathbf{W}_Q)^{\top} ([\mathbf{h}_{r_i} \| \Phi(t_i)]\mathbf{W}_K ) \right) \right)}{\sum_{(\Tilde{e}_j, r_j, t_j) \in \mathcal{N}_{e'}} \exp \left( \sigma \left( ([\mathbf{h}_{r_q} \| \Phi(t_q)] \mathbf{W}_Q)^{\top} ([\mathbf{h}_{r_j} \| \Phi(t_j)]\mathbf{W}_K ) \right) \right)}$. $\Phi$ denotes the functional time encoder proposed in \cite{DBLP:conf/iclr/XuRKKA20} and $\mathbf{W}_Q$, $\mathbf{W}_K$ are two weight matrices.

We report the experimental results of the ablation studies in Table \ref{tab: ablation}. From A1 to A3, we show that our concept modeling component helps to improve model performance, and it benefits from its double branch structure. From B1, we find that incorporating temporal information into the graph encoder is important for modeling TKGs. Besides, B2 to B4 show that in TKG few-shot OOG link prediction, it is not necessary to employ a complicated graph encoding structure. A possible reason is that we can only observe $K$ (1 or 3) associated quadruples for every unseen entity, and this forms a tiny neighborhood. Complicated structures, e.g., our time-aware attentional network, are unable to demonstrate their superiority in this case.
\begin{table*}[htbp]
    \centering
    \resizebox{0.7\columnwidth}{!}{
    \begin{tabular}{@{}lccc ccc ccc ccc @{}}
\toprule
        \textbf{Datasets} & \multicolumn{6}{c}{\textbf{ICEWS14-OOG}} & \multicolumn{6}{c}{\textbf{ICEWS18-OOG}}\\
\cmidrule(lr){2-7} \cmidrule(lr){8-13} 
& \multicolumn{2}{c}{MRR} & \multicolumn{2}{c}{H@1}  & \multicolumn{2}{c}{H@10}& \multicolumn{2}{c}{MRR} & \multicolumn{2}{c}{H@1}  & \multicolumn{2}{c}{H@10} \\

 \textbf{Model}& 1-S & 3-S & 1-S & 3-S  & 1-S & 3-S & 1-S & 3-S & 1-S & 3-S  & 1-S & 3-S \\
\midrule
        A1 & .267 & .302 & .195 & .220 &.407 & .462
        & .187 & .261 & .128 & .181 & .315 & .408
        
         \\
        A2 & .271 & .285 & .203 & .217 & .403 & .454
        & .188 & .265 & .129 & .187 & .316 & .411
        
         \\
        A3 & .276 & .306 & .206 & .235 & .401 & .471
        & .189 & .265 & .125 & .185 & .316 & .415
        
         \\
\midrule
        B1 & .243 & .256&.171 & .179&.361 & .402
        & .184 & .238 & .122 & .162 & .314 & .383
        
         \\
        B2  & .258 & .281&.181 & .196&.393 & .432
        & .185 & .240 & .119 & .165 & .316 & .388
        
         \\
        B3  & .249 & .278&.177 & .179&.389 & .438 
        & .183 & .242 & .116 & .166 & .314 & .395
        
       \\
       B4  & .263 & .284&.192 & .195&.400 & .450
        & .181 & .245 & .112 & .174 & .307 & .393
       
       \\
\midrule
        FILT 
        & \textbf{.278} & \textbf{.321}&\textbf{.208} & \textbf{.240}&\textbf{.410}&\textbf{.475}
        & \textbf{.191} & \textbf{.266}&\textbf{.129} & \textbf{.187}&\textbf{.316}&\textbf{.417}
        \\
\bottomrule
    \end{tabular}
    }
    \caption{Ablation studies of FILT on ICEWS14-OOG and ICEWS18-OOG. H@1/3/10 denote Hits@1/3/10, respectively. The best results are marked in bold.}\label{tab: ablation}
\end{table*}

\subsection{Further Analysis}
\paragraph{Cross shot analysis.}
We evaluate our trained 3-shot and 1-shot models with varying shots (1,3 or 5-shot) during meta-test. We observe in Table \ref{tab: cross shot} that for both trained models, the performance increases as the test shot size rises. This is due to the effectiveness of our time-aware graph encoder. It distinguishes the importance of different support quadruples and better incorporates graph information as the shot size increases. We also observe that when the test shot size is larger than 3, FILT trained with 3 shots performs better than it trained with 1 shot. This is because during 3-shot meta-training, we simulate that for every unseen entity, 3 support examples are observable, which helps the model to generalize to the cases where their shot sizes are larger than 1 during meta-test. Besides, test with random shots does not greatly affect our model performance, thus showing FILT's robustness.
\begin{table}[htbp]\centering
\resizebox{\columnwidth}{!}{
\begin{tabular}{lcccccccccccccccccc}
\toprule
\textbf{Datasets} & \multicolumn{6}{c}{\textbf{ICEWS14-OOG}} & \multicolumn{6}{c}{\textbf{ICEWS18-OOG}} & \multicolumn{6}{c}{\textbf{ICEWS0515-OOG}}\\
\midrule
& \multicolumn{3}{c}{\textbf{(Train) 1-shot}} & \multicolumn{3}{c}{\textbf{(Train) 3-shot}} & \multicolumn{3}{c}{\textbf{(Train) 1-shot}} & \multicolumn{3}{c}{\textbf{(Train) 3-shot}}& \multicolumn{3}{c}{\textbf{(Train) 1-shot}} & \multicolumn{3}{c}{\textbf{(Train) 3-shot}}\\
\cmidrule(lr){2-4} \cmidrule(lr){5-7} \cmidrule(lr){8-10} \cmidrule(lr){11-13} \cmidrule(lr){14-16} \cmidrule(lr){17-19}
 \textbf{Test Shots}& MRR & H@1 & H@10 & MRR & H@1 & H@10 & MRR & H@1 & H@10 & MRR & H@1 & H@10 & MRR & H@1 & H@10 & MRR & H@1 & H@10 \\
\midrule
1-shot & .278 & .208 & .410 & .265 & .195 & .386 & .191 & .129 & .316 & .178 & .117 & .305 & .273 & .201 & .405 & .258 & .184 & .399\\
3-shot & .293 & .212 & .452 & .321 & .240 & .475 & .232 & .158 & .381 & .266 & .187 & .417 & .331 & .254 & .482 & .370 & .299 & .516\\
5-shot & .297 & .212 & .467 & .322 & .231 & .503 & .256 & .183 & .400 & .289 & .206 & .449 & .351 & .275 & .499 & .394 & .317 & .553\\
R-shot & .283 & .203 & .440 & .299 & .214 & .462 & .224 & .154 & .364 & .242 & .167 & .390 & .315 & .240 & .460 & .337 & .262 & .490\\
\bottomrule
\end{tabular}
}
\caption{Cross shot analysis results. R-shot denotes the setting that we randomly sample 1, 3 or 5 support quadruples for every unseen entity during meta-test. H@1/3/10 denote Hits@1/3/10, respectively.} \label{tab: cross shot}
\end{table}

\begin{figure}[htbp]
\centering
\includegraphics[width=\columnwidth]{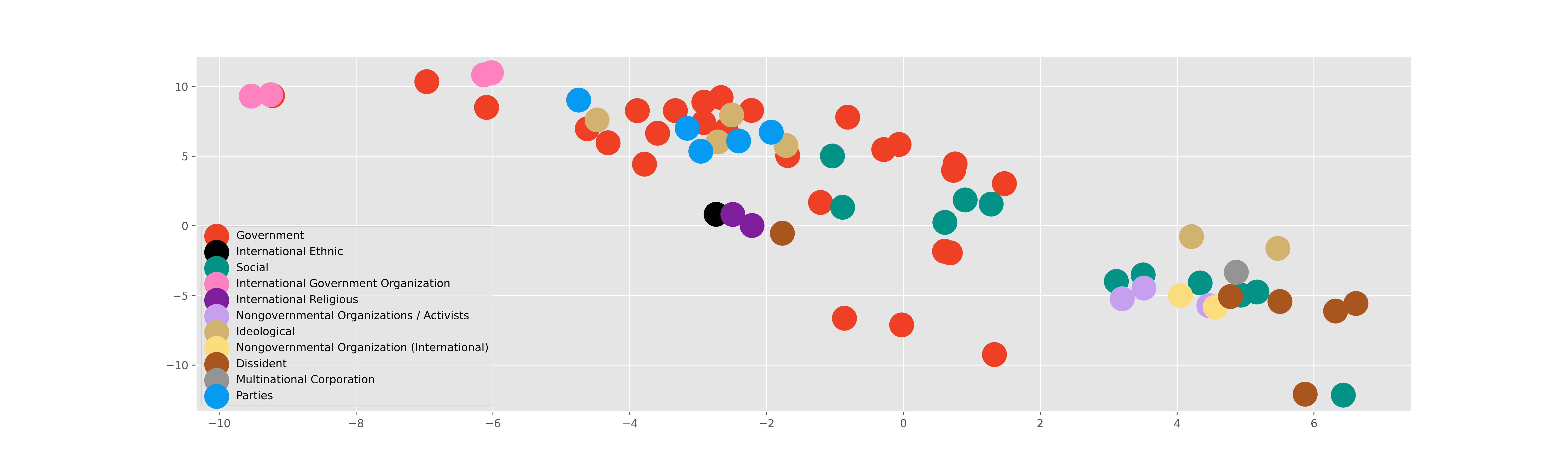}
\caption{\label{fig: visualization} Visualization of learned concept representations on 3-shot ICEWS18-OOG.}
\end{figure}

\paragraph{Visualization of concept representations.}
We plot the trained concept representations of the 3-shot model on ICEWS18-OOG with t-SNE \cite{van2008visualizing}. The entity concepts in the ICEWS database are hierarchical. For example, under the concept \textit{Government}, there exist other concepts, e.g., \textit{Foreign Ministry}. We only create labels for the first hierarchy concepts and assign other concepts belonging to them with the same label. From Figure \ref{fig: visualization}, we can observe that the concepts bearing the same label tend to form a cluster, and the clusters having similar semantic meanings tend to be close to each other, e.g., the clusters of \textit{Parties} and \textit{Government}. This demonstrates that our concept modeling component learns the semantics of entity concepts, which helps to improve inductive learning for unseen new entities. We present three case studies in Appendix \ref{app: case study}.

\section{Conclusion}
We propose a new task: temporal knowledge graph (TKG) few-shot out-of-graph (OOG) link prediction, aiming to introduce the inductive entity representation learning problem into TKGs. We develop a model that focuses on the \textbf{f}ew-shot \textbf{i}nductive \textbf{l}earning on \textbf{T}KGs (FILT). Given only few edges associated to each newly-emerged entity, FILT employs a meta-learning framework that enables inductive knowledge transfer from seen entities to new unseen entities. FILT uses a time-aware graph encoder to learn the contextualized representations of unseen entities, which shows stronger performance as the shot size increases. It also utilizes the external entity concept information specified in the temporal knowledge bases. We propose three new datasets for TKG few-shot OOG link prediction and compare FILT with several baselines. Experimental results show that learning concept-aware information improves inductive learning for emerging entities. 
In the future, we would like to generalize rule-based knowledge graph reasoning methods to the TKG inductive learning scenario. Another direction is to combine future link prediction with our proposed TKG few-shot OOG link prediction task since our task currently does not support link forecasting.

\bibliography{sample}
\bibliographystyle{plainnat}

\appendix
\section{Long-Tail Distribution of Entities in Temporal Knowledge Bases}
\label{app: long tail}
Figure \ref{app: long tail} illustrates the entity occurrence of ICEWS14, ICEWS18 and ICEWS05-15 databases. We find that most entities occur for only a few times.
\begin{figure}[htbp]
\centering
\includegraphics[width=.3\textwidth]{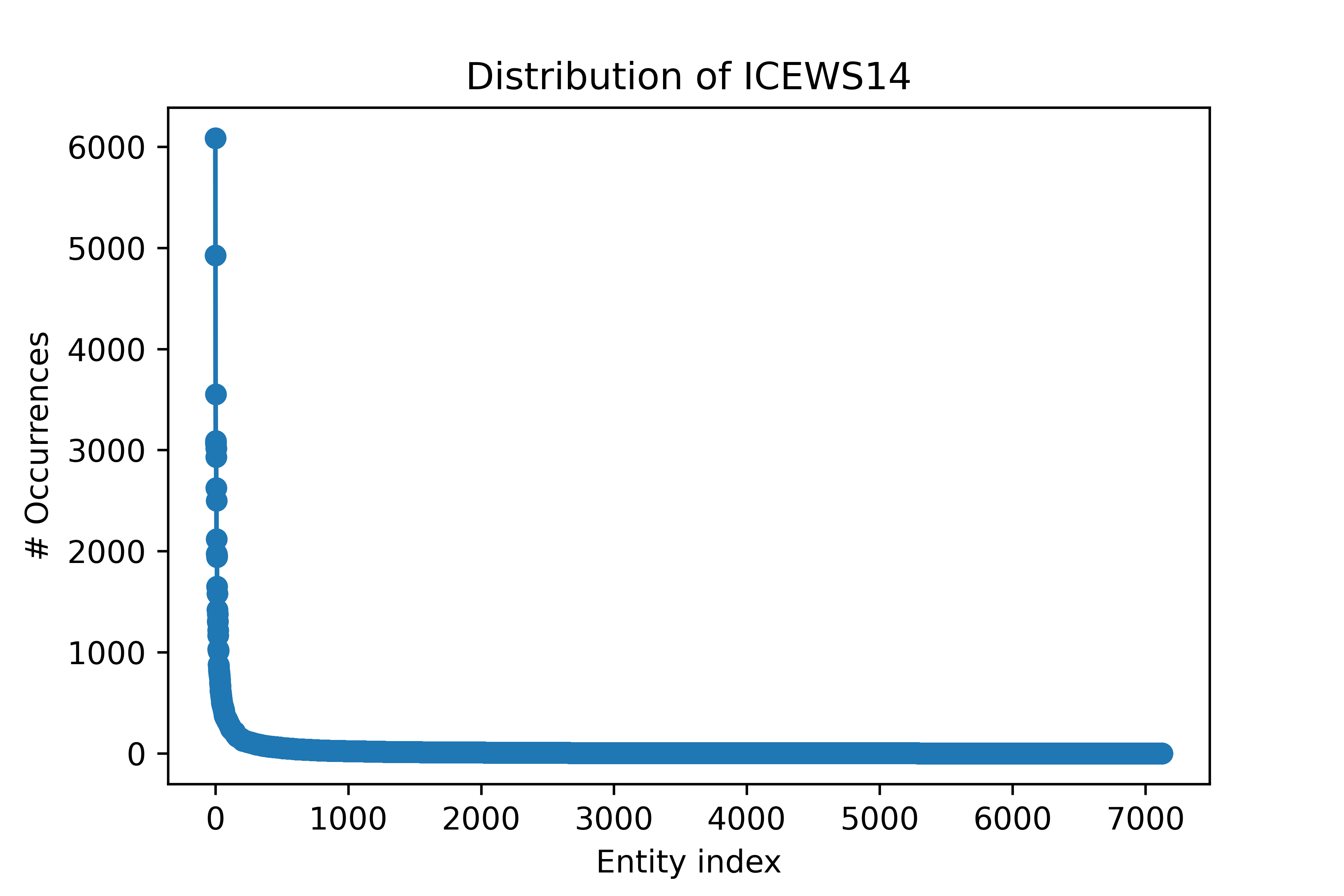}\hfill
\includegraphics[width=.3\textwidth]{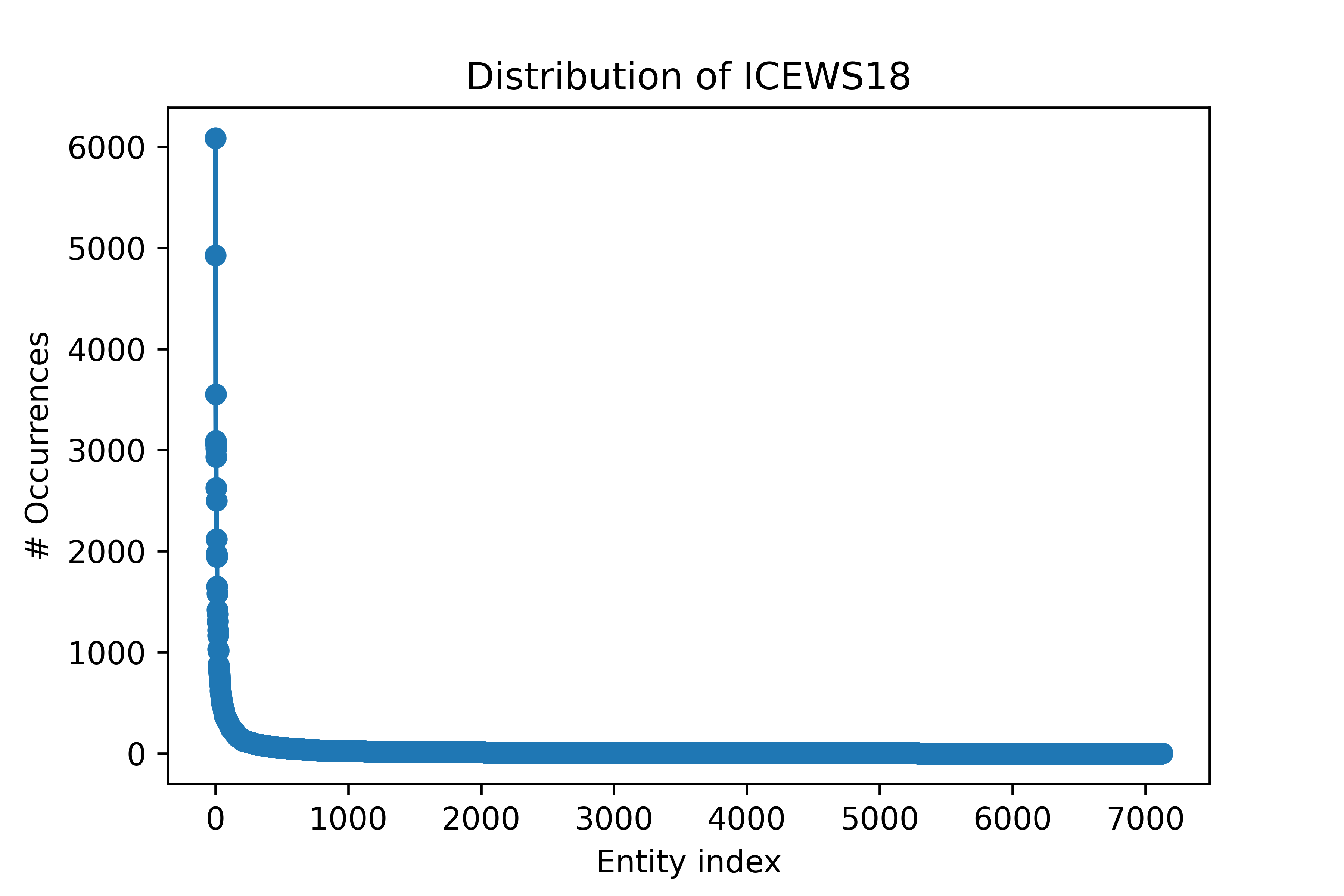}\hfill
\includegraphics[width=.3\textwidth]{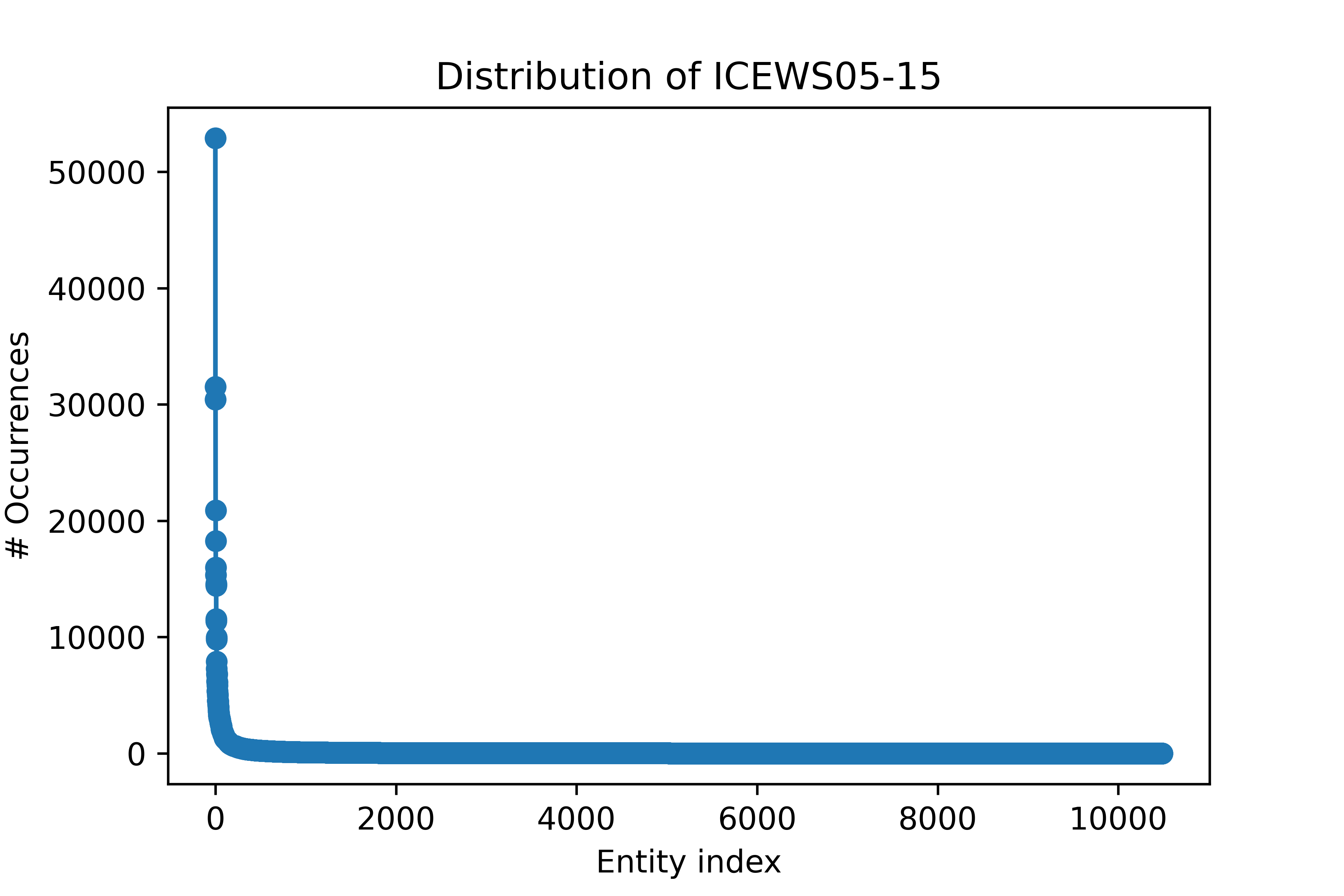}
\caption{Entity occurrence of ICEWS14, ICEWS18 and ICEWS05-15 databases.}
\label{fig: long tail}

\end{figure}
\section{Evaluation Metrics}
\label{app: eval metrics}
We use two evaluation metrics for our experiments, i.e., mean reciprocal rank (MRR) and Hits@1/3/10. For every link prediction query, we compute the rank $\psi$ of the ground truth missing entity. MRR is defined as: $\frac{1}{\sum_{e' \in \mathcal{E}'_{\text{meta-test}}} |\mathcal{Q}_{e'}|}\sum_{e' \in \mathcal{E}'_{\text{meta-test}}} \sum_{q^+ \in \mathcal{Q}_{e'}} \frac{1}{\psi}$. Hits@1/3/10 denote the proportions of the predicted links where ground truth missing entities are ranked as top 1, top3, top10, respectively.

\section{Implementation Details}
We implement all the experiments with PyTorch \cite{DBLP:conf/nips/PaszkeGMLBCKLGA19} on a single NVIDIA Tesla T4. We search hyperparameters following Table \ref{tab: hyperparameter search}. For each dataset, we do 108 trials to try different hyperparameter settings. We run 15000 batches for each trail and compare their meta-validation results. We choose the setting leading to the best meta-validation result and take it as the best hyperparameter setting. We report the best hyperparameter setting in Table \ref{tab: best param}. Every result of our model is the average result of five runs. For the models leading to the results reported in Table \ref{tab: main results}, we provide their meta-validation results in Table \ref{tab: valid results}. We also specify their GPU memory usage (Table \ref{tab: memory}) and number of parameters (Table \ref{tab: num of param}). For different datasets, we use different numbers of unseen entities $N$ in each meta-training task $T$. We set $N=100$ for ICEWS14-OOG and ICEWS0515-OOG, $N=200$ for ICEWS18-OOG. We sample 32 negative samples for every positive sample.
\begin{table}[!htb]
    \begin{minipage}{.435\linewidth}
      \centering
      \resizebox{\linewidth}{!}{
        \begin{tabular}{ll}
            \toprule 
       \textbf{Hyperparameter} & \textbf{Search Space} \\
       \midrule 
       Embedding Size & \{50, 100, 200\}     \\
       \# Aggregation Step &  \{1, 2\}   \\
       Activation Function &  \{Tanh, ReLU, LeakyReLU\}\\
       Dropout & \{0.2, 0.3, 0.5\}\\
       $\lambda$ & \{0.2, 0.4\}\\
       
      \bottomrule 
        \end{tabular}
        }
        \caption{Hyperparameter searching strategy.}
        \label{tab: hyperparameter search}
    \end{minipage}%
    \begin{minipage}{.565\linewidth}
      \centering
        \resizebox{\linewidth}{!}{
        \begin{tabular}{lccc} 
      \toprule 
     \multicolumn{1}{l}{\textbf{Datasets}} & \multicolumn{1}{c}{\textbf{ICEWS14-OOG}} & \multicolumn{1}{c}{\textbf{ICEWS18-OOG}} &\multicolumn{1}{c}{\textbf{ICEWS0515-OOG}} \\
      \midrule 
      \textbf{Hyperparameter} &   &  &   \\
       \midrule 
       Embedding Size & 100     &  100&  100 \\
       \# Aggregation Step &  1   &  1&   1 \\
       Activation Function &  LeakyReLU&  LeakyReLU & LeakyReLU\\
       Dropout & 0.3& 0.3 & 0.3\\
       $\lambda$ & 0.2 & 0.4 & 0.4\\
       
      \bottomrule 
    \end{tabular} 
    }
    \caption{Best hyperparameter settings.}
    \label{tab: best param}
    \end{minipage} 
\end{table}
\begin{table*}[htbp]
    \centering
    \resizebox{\columnwidth}{!}{
    \begin{tabular}{@{}lcccc cccc cccc cccc cccc cccc@{}}
\toprule
        \textbf{Datasets} & \multicolumn{8}{c}{\textbf{ICEWS14-OOG}} & \multicolumn{8}{c}{\textbf{ICEWS18-OOG}} & \multicolumn{8}{c}{\textbf{ICEWS0515-OOG}}\\
\cmidrule(lr){2-9} \cmidrule(lr){10-17} \cmidrule(lr){17-25}
& \multicolumn{2}{c}{MRR} & \multicolumn{2}{c}{H@1} & \multicolumn{2}{c}{H@3} & \multicolumn{2}{c}{H@10}& \multicolumn{2}{c}{MRR} & \multicolumn{2}{c}{H@1} & \multicolumn{2}{c}{H@3} & \multicolumn{2}{c}{H@10} & \multicolumn{2}{c}{MRR} & \multicolumn{2}{c}{H@1} & \multicolumn{2}{c}{H@3} & \multicolumn{2}{c}{H@10}\\

 \textbf{Model}& 1-S & 3-S & 1-S & 3-S & 1-S & 3-S & 1-S & 3-S & 1-S & 3-S & 1-S & 3-S & 1-S & 3-S & 1-S & 3-S & 1-S & 3-S & 1-S & 3-S & 1-S & 3-S & 1-S & 3-S\\
\midrule
        FILT 
        & .251 & .354&.171 & .271&.285 &.389 &.410&.511
        & .187 & .242&.127 & .163&.204 & .264&.308&.406
        & .232 & .316&.163 & .229&.247 & .350&.378&.491
        \\
\bottomrule
    \end{tabular}
    }
    \caption{TKG 1-shot and 3-shot OOG link prediction results on the meta-validation set. Evaluation metrics are filtered MRR and Hits@1/3/10 (H@1/3/10).}\label{tab: valid results}
\end{table*}

\begin{table}[!htb]
    \begin{minipage}{.5\linewidth}
      \centering
      \resizebox{\linewidth}{!}{
      \begin{tabular}{lcccccc} 
      \toprule 
     \multicolumn{1}{l}{\textbf{Datasets}} & \multicolumn{2}{c}{\textbf{ICEWS14-OOG}} & \multicolumn{2}{c}{\textbf{ICEWS18-OOG}} &\multicolumn{2}{c}{\textbf{ICEWS0515-OOG}} \\
      \cmidrule(lr){2-3}\cmidrule(lr){4-5}\cmidrule(lr){6-7}
       & \multicolumn{2}{c}{GPU Memory}  & \multicolumn{2}{c}{GPU Memory} & \multicolumn{2}{c}{GPU Memory}  \\
      \textbf{Model} & 1-S & 3-S & 1-S & 3-S & 1-S & 3-S\\
       \midrule
       FILT & 1493MB & 1466MB & 1871MB & 1841MB & 1557MB & 1541MB \\
       
      \bottomrule 
    \end{tabular}
        }
        \caption{GPU memory usage.}
    \label{tab: memory}
    \end{minipage}%
    \begin{minipage}{.5\linewidth}
      \centering
        \resizebox{\linewidth}{!}{
        \begin{tabular}{lcccccc} 
      \toprule 
     \multicolumn{1}{l}{\textbf{Datasets}} & \multicolumn{2}{c}{\textbf{ICEWS14-OOG}} & \multicolumn{2}{c}{\textbf{ICEWS18-OOG}} &\multicolumn{2}{c}{\textbf{ICEWS0515-OOG}} \\
      \cmidrule(lr){2-3}\cmidrule(lr){4-5}\cmidrule(lr){6-7}
       & \multicolumn{2}{c}{\# Param}  & \multicolumn{2}{c}{\# Param} & \multicolumn{2}{c}{\# Param} \\
      \textbf{Model} & 1-S & 3-S & 1-S & 3-S & 1-S & 3-S\\
       \midrule
       FILT & 2966303 & 2966303 & 4567203 & 4567203 & 3310703 & 3310703 \\
       
      \bottomrule 
    \end{tabular} 
        
    }
    \caption{Number of parameters.}
    \label{tab: num of param}
    \end{minipage} 
\end{table}

For baseline methods, except MEAN, we use their official implementations, i.e., ComplEx\footnote{https://github.com/ttrouill/complex}, BiQUE\footnote{https://github.com/guojiapub/BiQUE}, TNTComplEx\footnote{https://github.com/facebookresearch/tkbc}, TeLM\footnote{https://github.com/soledad921/TeLM}, TeRo\footnote{https://github.com/soledad921/ATISE}, LAN\footnote{https://github.com/wangpf3/LAN}, GEN\footnote{https://github.com/JinheonBaek/GEN}. We use the MEAN implementation provided in the LAN repository. We use default hyperparameters of TKGC methods for ICEWS datasets. For other methods, we keep their embedding size the same as FILT's. We keep other hyperparameters of them as their default settings.

\section{Further Discussion of TKG Few-Shot OOG Link Prediction}
\label{app: oog link prediction}
\begin{figure}[htbp]
\centering
\includegraphics[width=0.8\columnwidth]{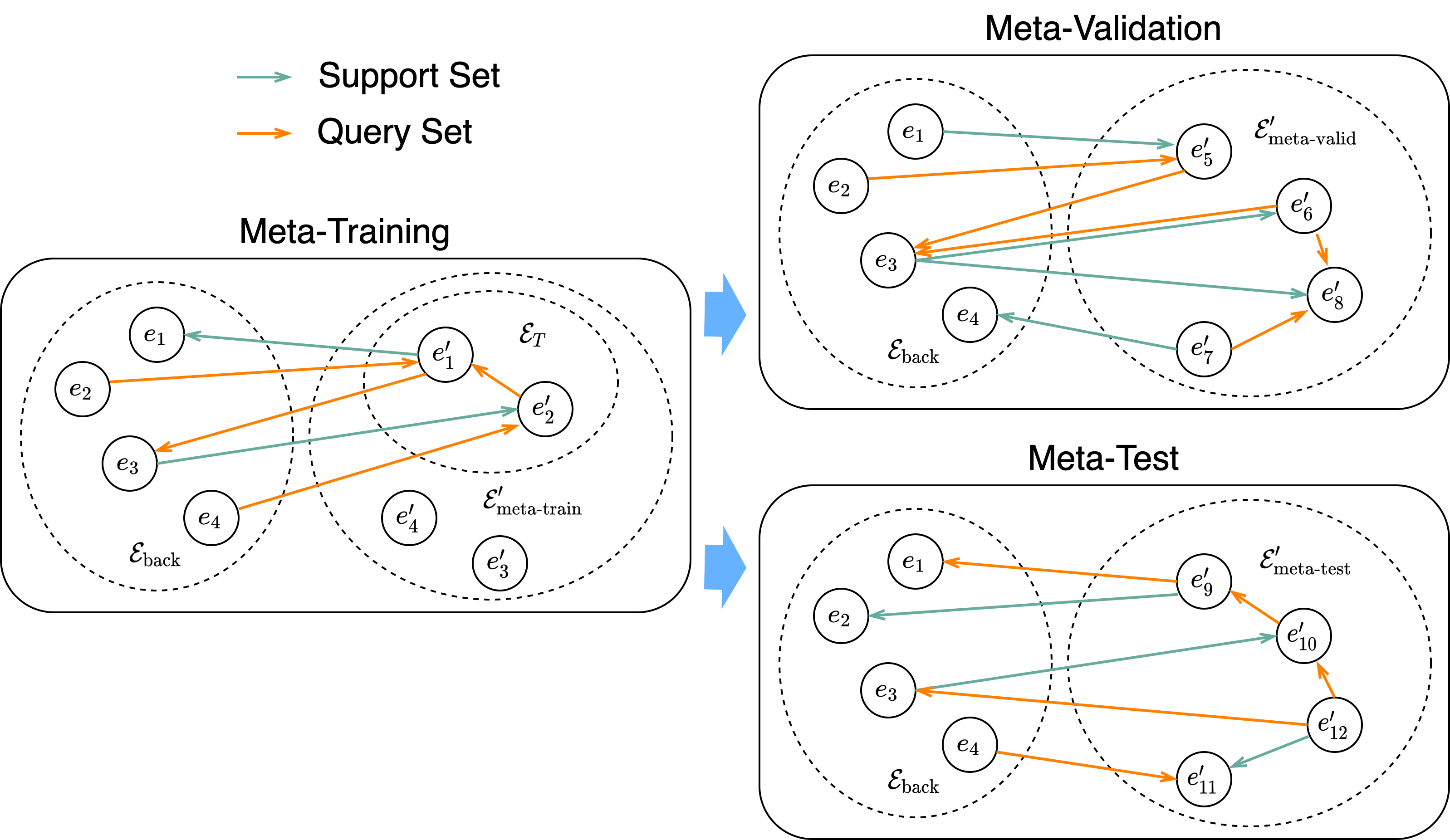}
\caption{\label{fig: meta-learning} Illustration of the meta-learning framework formulated from the TKG few-shot OOG link prediction task.}
\end{figure}

Figure \ref{fig: meta-learning} illustrates how we formulate TKG few-shot OOG link prediction into a meta-learning problem with an example. Green edges correspond to the support quadruples and orange edges correspond to the query quadruples (timestamps and relations are omitted for brevity). The meta-training process consists of a number of meta-training tasks. During each meta-training task $T$, $N$ unseen entities from $\mathcal{E}'_\text{meta-train}$ are randomly sampled. In Figure \ref{fig: meta-learning}, $e'_1, e'_2 \in \mathcal{E}'_\text{meta-train}$ are sampled in task $T$. For each sampled unseen entity, $K$ ($K = 1$ in Figure \ref{fig: meta-learning}) quadruples from all the quadruples containing itself are sampled to form its support set. The rest form its query set. During meta-validation, all the unseen entities ($e'_5, e'_6, e'_7, e'_8$) from $\mathcal{E}'_\text{meta-valid}$ are treated as appearing simultaneously, which also applies to meta-test and the unseen entities ($e'_9, e'_{10}, e'_{11}, e'_{12}$) from $\mathcal{E}'_\text{meta-test}$.

\paragraph{TKG few-shot OOG link prediction vs. TKG completion.}
For the existing TKGC benchmark datasets, e.g., ICEWS14\footnote{https://github.com/BorealisAI/de-simple}, there exist a number of entities that only appear in the test sets (or the validation sets) and are unseen in their training sets. Evaluating on the links concerning these unseen entities coincides to the evaluation setting of TKG few-shot OOG link prediction. However, in our proposed task, we focus on the unseen entities that are long-tail, and we also introduce a realistic setting that each unseen entity is coupled with $K$ support quadruples containing itself, while in traditional TKGC benchmark datasets the unseen entities are not guaranteed to be long-tail and no associated edge is given for learning the inductive representations of them. The aim of TKG few-shot OOG link prediction is to ask the TKG reasoning models to learn strong representations of the unseen entities inductively from extracting the information from the provided $K$ support quadruples, which corresponds to the realistic situation where every newly-emerged entity is often coupled with a small number of associated edges.

\section{Ablation Study Details}
We present the detailed equations of graph encoder variants (B1-B3 in Table \ref{tab: ablation}, B4 already presented). In B1, RGCN computes the unseen entity $e'$'s representation as:
\begin{equation}
    \mathbf{h}_{(e', t_q)} = \frac{1}{|\mathcal{N}_{e'}|}\sum_{(\Tilde{e}_i, r_i, t_i) \in \mathcal{N}_{e'}}   \mathbf{W}_{r_i}(\mathbf{h}_{\Tilde{e}_i}),
\end{equation}
where $\mathbf{W}_{r_i}$ is a weight matrix modeling $r_i$. In B2, Time2Vec computes $e'$'s representation as:
\begin{equation}
\label{eq: time2vec}
    \textbf{h}_{(e', t_q)} = \frac{1}{|\mathcal{N}_{e'}|} \sum_{(\Tilde{e}_i, r_i, t_i) \in \mathcal{N}_{e'}} \textbf{W}_g (\textbf{h}_{(\Tilde{e}_i, t_i)} \| \textbf{h}_{r_i}),
\end{equation}
where $\textbf{h}_{(\Tilde{e}_i, t_i)}$ is defined as:
\begin{equation}
\begin{aligned}
        \textbf{h}_{(\Tilde{e}_i, t_i)} &= f(\textbf{h}_{\Tilde{e}_i} \| \Phi(t_i)), \\
        \Phi(t_i)[j] &= \left\{ \begin{multlined}  \quad \omega_j t_i + \varphi_j, \quad \text{if}\ j=0, \\ 
        \sin (\omega_j t_i + \varphi_j), \quad \text{if}\  1 \leq j \leq d_t.  \end{multlined} \right.
\end{aligned}
\end{equation}
$f$ denotes a layer of feed forward neural network. $\Phi(t_i)[j]$ denotes the $j$th component of $t_i$'s time representation $\Phi(t_i)$. $d_t$ is the dimension size of time representations. $\omega_j$ and $\varphi_j$ represent the trainable frequency and phase parameters, respectively.
In B3, we use the same aggregation function \ref{eq: time2vec} as in Time2vec, however, we use another form of time encoder to encode time information:
\begin{equation}
\begin{aligned}
        \textbf{h}_{(\Tilde{e}_i, t_i)} &= f(\textbf{h}_{\Tilde{e}_i} \| \Phi(t_i)), \\
        \Phi(t_i) &= \sqrt{\frac{1}{d_t}}[\cos(\omega_1t_i+\varphi_1), \dots, \cos(\omega_{d_t}t_i+ \varphi_{d_t}))],
\end{aligned}
\end{equation}
where $\omega_1 \dots \omega_{d_t}$ and $\phi_1 \dots \phi_{d_t}$ are trainable parameters.
\section{Concept Extraction of ICEWS Database}
\label{app: concept extract}
We take the sectors of ICEWS entities as their concepts. The sector classification can be found on the ICEWS official website\footnote{https://dataverse.harvard.edu/dataset.xhtml?persistentId=doi:10.7910/DVN/28118}. ICEWS sectors have hierarchies. We do not consider hierarchies and consider each sector as individual. For example, the sector \textit{Foreign Ministry} belongs to the sector \textit{Government}. We learn their representations separately.

A number of entities in the ICEWS coded event data are not labeled with any sector. Some of them are regions, e.g., \textit{North Korea}. We create a new sector named \textit{Region} for them. For other entities, we find their affiliations and pick out their sectors. We then choose from their affiliations' sectors the most suitable ones and label these entities. For example, \textit{European Parliament} has no associated sector in the ICEWS coded event data. We find its affiliation \textit{European Union}. \textit{European Union} is assigned a sector \textit{Regional Diplomatic IGOs}. We take \textit{Regional Diplomatic IGOs} as \textit{European Parliament}'s sector and it is taken as a concept in our meta-learning process.

\section{Case Study of Learned Concept Representations}
\label{app: case study}
We further find three cases to show that our learned concept representations capture the semantic meaning of concepts, which helps to embed unseen entities inductively. We resize the visualization in Figure \ref{fig: visualization} and label several concepts close to each other.
\begin{figure}[htbp]
\centering
\includegraphics[width=\columnwidth]{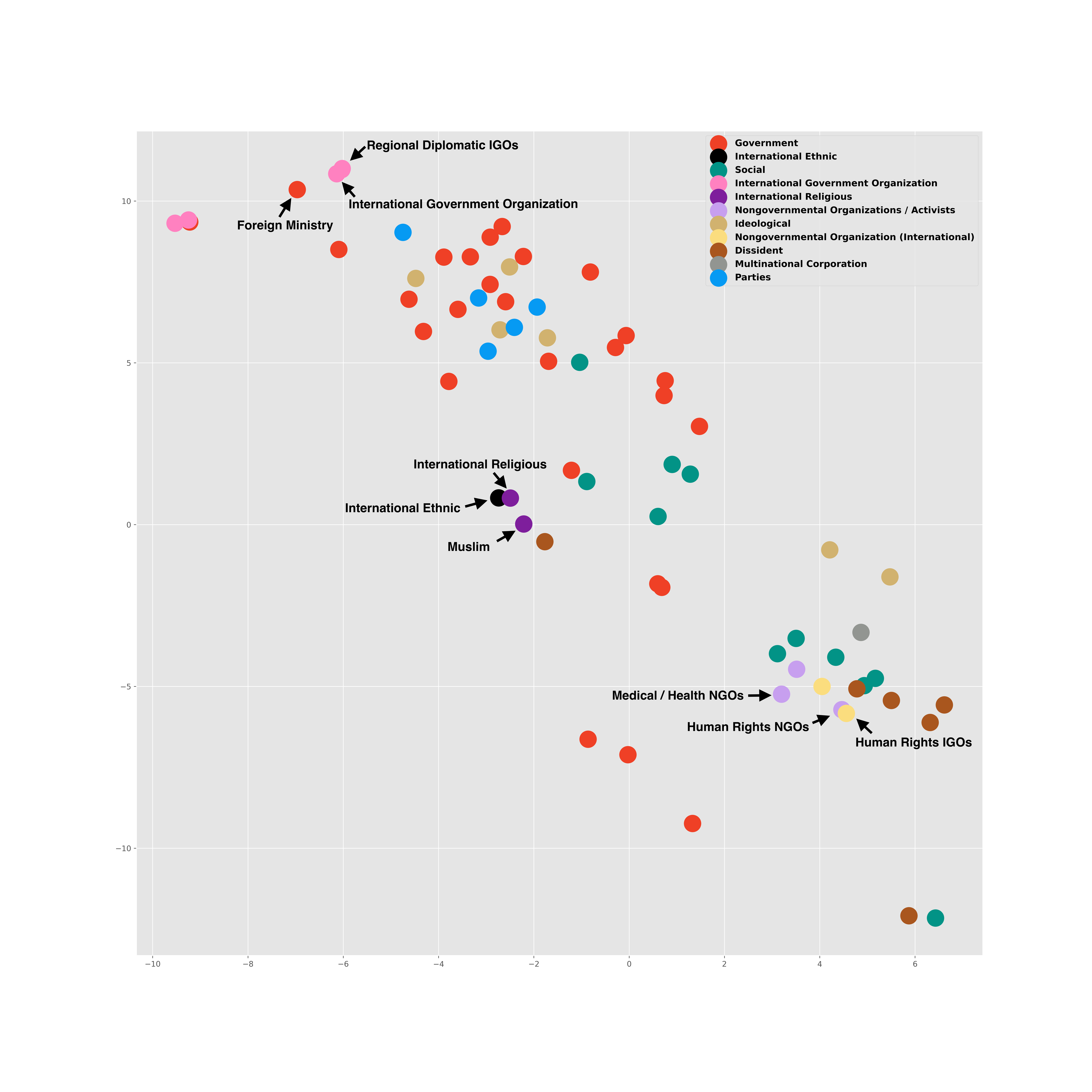}
\caption{\label{fig: visualization appendix} Resized visualization of learned concept representations on 3-shot ICEWS18-OOG.}
\end{figure}

The first case is about the concepts \textit{Foreign Ministry}, \textit{International Government Organization} and \textit{Regional Diplomatic IGOs}, where \textit{IGO} stands for international government organization. From human intuition, \textit{Foreign Ministry} is closely related to international interactions. Similarly, \textit{international Government Organization} and \textit{Regional Diplomatic IGOs} also possess the same semantics.

The second case is about the concepts \textit{International Ethnic}, \textit{International Religious} and \textit{Muslim}. \textit{Muslim} stands for not only a religion but also an ethnicity, therefore, it is close to both \textit{International Religious} and \textit{International Ethnic}.

The third case is about the concepts \textit{Medical / Health NGOs}, \textit{Human Rights NGOs} and \textit{Human Rights IGOs}, where \textit{NGO} stands for nongovernmental organizations. We can observe that \textit{Human Rights NGOs} and \textit{Human Rights IGOs} are extremely close to each other. Since protecting human rights is normally concerned with providing medical aid, they are also close to \textit{Medical / Health NGOs}.

\section{Dataset Construction Process}
\label{app: data construct}
\begin{itemize}[label={}]
\item{1. }We take ICEWS14\footnote{https://github.com/BorealisAI/de-simple}, ICEWS18\footnote{https://github.com/INK-USC/RE-Net} and ICEWS05-15\footnote{https://github.com/mniepert/mmkb/tree/master/TemporalKGs} as the databases for dataset construction.
\item{2. }We set the upper and lower thresholds for entity frequencies. We do not want the upper threshold to be large since in real-world scenarios, newly-emerged entities normally are only coupled with very few edges. We also do not want the lower threshold to be too small since we want to include enough test examples. We set the upper and lower threshold to 10 and 25 for every dataset.
\item{3. }We pick out the entities whose frequencies are between thresholds and sample half of them as the total unseen entities $\mathcal{E}'$ (following \cite{DBLP:conf/nips/BaekLH20}). We take the quadruples without any unseen entity as the background graph $\mathcal{G}_{\text{back}}$.
\item{4. }We split the unseen entities as meta-training $\mathcal{E}'_{\text{meta-train}}$, meta-validation $\mathcal{E}'_{\text{meta-valid}}$ and meta-test $\mathcal{E}'_{\text{meta-test}}$ entities. $|\mathcal{E}'_{\text{meta-train}}|:|\mathcal{E}'_{\text{meta-valid}}|:|\mathcal{E}'_{\text{meta-test}}| \approx 8:1:1$. Their associated quadruples form the corresponding meta-learning sets.
\end{itemize}
\end{document}